\pdfoutput=1
\documentclass[11pt]{article}
\usepackage[]{acl}
\usepackage{times}
\usepackage{latexsym}
\usepackage{fnpct}
\usepackage{stmaryrd}
\usepackage{amsmath}
\usepackage{multirow}
\usepackage{bbm}
\usepackage{graphicx}
\usepackage{amssymb}
\usepackage{booktabs}
\usepackage{adjustbox}
\usepackage{xspace}
\usepackage{algpseudocode}
\usepackage{algorithm}
\usepackage{graphicx}
\usepackage{caption}
\usepackage{tikz}
\usepackage{subcaption}
\usepackage{hyperref}
\usepackage{colortbl}

\usepackage{xcolor}

\usepackage[T1]{fontenc}
\usepackage[utf8]{inputenc}

\usepackage{microtype}

\title{The Decades Progress on Code-Switching Research in NLP: \\A Systematic Survey on Trends and Challenges}

\author{Genta Indra Winata$^1$, Alham Fikri Aji$^2$, Zheng-Xin Yong$^3$, Thamar Solorio$^{1}\thanks{\hspace{0.2cm}The work was done while at Bloomberg.}$ \\
  $^1$Bloomberg \quad $^2$MBZUAI \quad $^3$Brown University \\
  \texttt{\small gwinata@bloomberg.net, alham.fikri@mbzuai.ac.ae, contact.yong@brown.edu}}

\begin{document}
\maketitle
\begin{abstract}
% Code-Switching has been a common phenomenon in written text and conversation that has been studied over decades by the natural language processing (NLP) research community.
Code-Switching, a common phenomenon in written text and conversation, has been studied over decades by the natural language processing (NLP) research community. Initially, code-switching is intensively explored by leveraging linguistic theories and, currently, more machine-learning oriented approaches to develop models. We introduce a comprehensive systematic survey on code-switching research in natural language processing to understand the progress of the past decades and conceptualize the challenges and tasks on the code-switching topic. Finally, we summarize the trends and findings and conclude with a discussion for future direction and open questions for further investigation.
\end{abstract}

\section{Introduction}
\begin{figure}[!t]
    \centering
    \begin{subfigure}{0.95\linewidth}
		\centering
		\includegraphics[width=\linewidth]{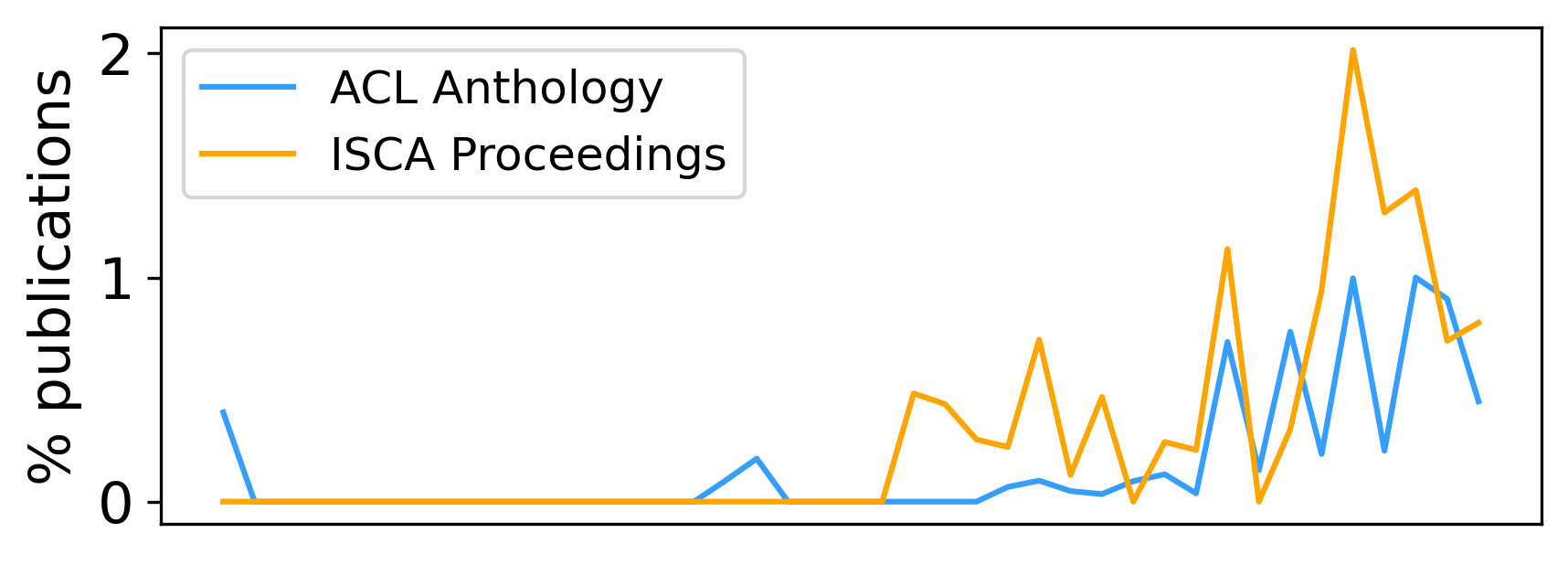}   
	\end{subfigure}
	\begin{subfigure}{0.95\linewidth}
		\centering
		\includegraphics[width=\linewidth]{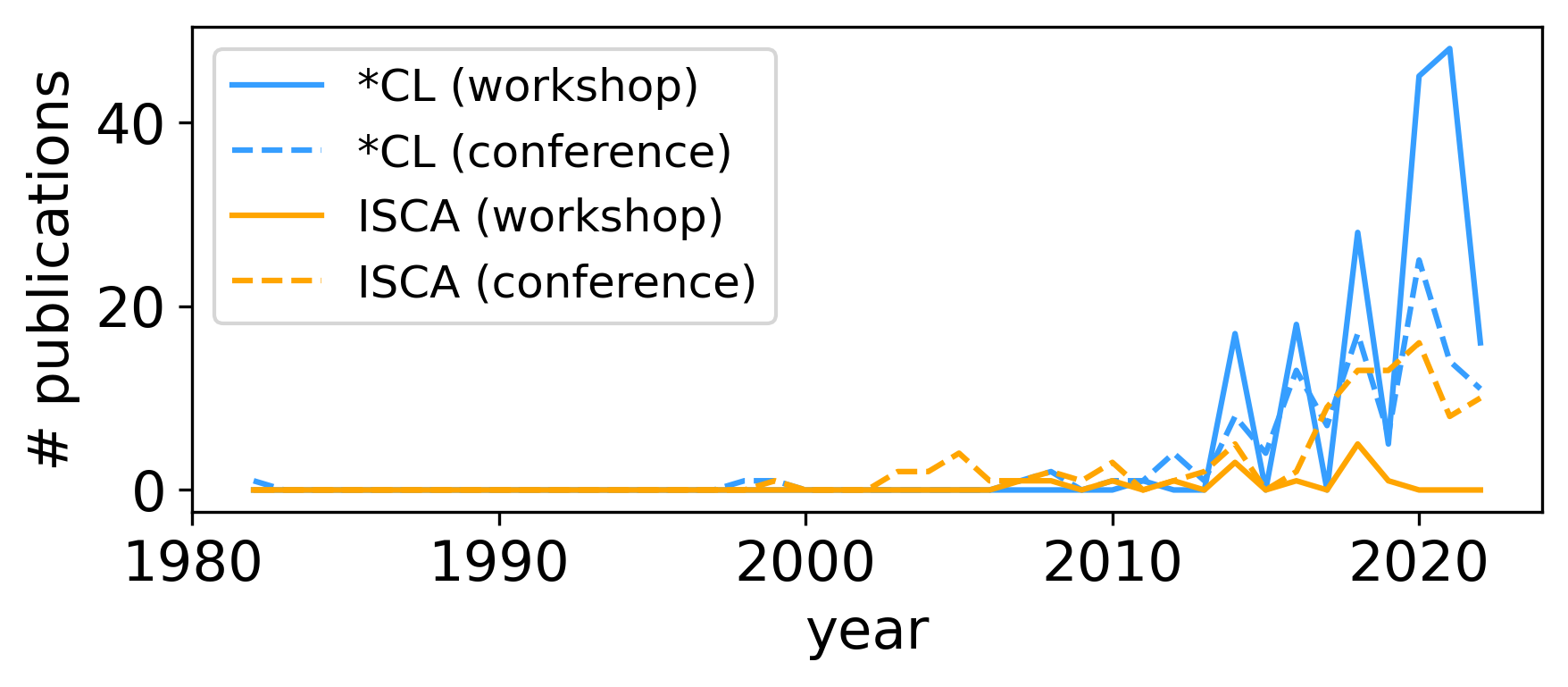} 
	\end{subfigure}
	\caption{Number of publications over time in *CL and ISCA venues. We collect the papers on October 2022. \textbf{Top:} Relative to all *CL and ISCA papers. \textbf{Bottom:} absolute number, broken down into conferences vs workshops. It does not include papers published after. The graphs do not show the number of publications published in journals and symposiums.}
 % \customtodo{compare with number of ACL submission}}
	\label{fig:publications}
\end{figure}

Code-Switching is the linguistic phenomenon where multilingual speakers use more than one language in the same conversation~\cite{poplack1978syntactic}. The fragment of the worldwide population that can be considered multilingual, i.e., speaks more than one language, far outnumbers monolingual speakers~\cite{tucker2001global,winata2021multilingual}. This alone makes a compelling argument for developing NLP technology that can successfully process code-switched (CSW) data. However, it was not until the last couple of years that CSW-related research became more popular~\cite{sitaram2019survey,jose2020survey,dougruoz2021survey}, and this increased interest has been motivated to a large extent by: 1) The need to process social media data. Before the proliferation of social media platforms, it was more common to observe code-switching in spoken language and not so much in written language. This is not the case anymore, as multilingual users tend to combine the languages they speak on social media; 2) The increasing release of voice-operated devices. Now that smart assistants are becoming more and more accessible, we have started to realize that assuming users will interact with NLP technology as monolingual speakers is very restrictive and does not fulfill the needs of real-world users. Multilingual speakers also prefer to interact with machines in a CSW manner \cite{bawa2020multilingual}. We show quantitative evidence of the upward trend for CSW-related research in Figure~\ref{fig:publications}. 

%We reviewed 414 papers from ACL and ISCA proceedings

\begin{table*}[!t]
    \centering
    \resizebox{\linewidth}{!}{
        \begin{tabular}{@{}ll@{}}
        \toprule
            \textbf{Category} & \textbf{Options} \\ \midrule
            Languages & Bilingual, Trilingual, 4+ \\
            % \quad Bilingual (Coarse) & xx-en, xx-yy \\ 
            % \quad XX-en (Fine) & European - English, South Asian - English, Middle Eastern - English, \\ 
            % & East Asian - English, South East Asian - English, African - English \\ 
            % \quad Trilingual (Coarse) & x-en, x-y \\ 
            % \quad Trilingual (Fine) &  \\ \midrule
            Venues & Conference, Workshop, Symposium, Book\\
            Papers & Theory / Linguistics, Empirical, Analysis, Position/Opinion/Survey, Metric, Corpus, Shared Task, Demo \\
            Datasets & Social Media, Speech (Recording), Transcription, News, Dialogue, Books, Government Document, Treebank \\
            Methods & Rule/Linguistic Constraint, Statistical Model, Neural Network, Pre-trained Model \\
            Tasks & \textbf{Text: } Topic Modeling, Semantic Parsing, Dependency Parsing, Sentiment Analysis, Emotion Detection, \\ & Abusive Language Detection, Sarcasm Detection, Humor Detection, Humor Generation, Dialogue State Tracking, \\ & Text Generation, Natural Language Understanding, Named Entity Recognition, Part-of-Speech Tagging, 
            \\ & Natural Language Entailment, Language Modeling, Regression, Language Identification,  Machine Translation, \\ &  Text Normalization, Micro-Dialect Identification,  
            Question Answering, Summarization \\
            & \textbf{Speech: } Acoustic Modeling, Speech Recognition, Text-to-Speech, Speech Synthesis \\
        \bottomrule
        \end{tabular}
    }
    \caption{Categories in the annotation scheme.}
    \label{tab:category}
\end{table*}

%The research on code-switching has been recently more attractive to researchers who study multilingual, and it is reflected by the number of publication on code-switching. Figure~\ref{fig:publications} shows a positive increment of the number of publications over the recent years. Thus, the field has been broadened and we are interested to understand more the latest trend on code-switching and study the significant factors they positively influence the growth of the field. 
% There are few NLP survey papers exist, covering specific topics in code-switching~\cite{sitaram2019survey,jose2020survey,dougruoz2021survey}. \citet{sitaram2019survey} investigates the methodologies on NLP for code-switching and \citet{jose2020survey} focuses on code-switching datasets. \citet{dougruoz2021survey} studies the linguistic perspective of code-switching research. However, they do not provide a full comprehensive study on the code-switching papers on the open proceedings and many of the previous work in the past decades are not included in the survey. In addition, we notice that a systematic process of annotating papers is required to have a holistic review of the field, and thus, we can understand how the trend of the field evolves through time.

In this paper, we present the first large-scale comprehensive survey on CSW NLP research in a structured manner by collecting more than 400 papers published on open repositories, such as the ACL Anthology and ISCA proceedings (see \textsection \ref{sec:annotation}). We manually coded these papers to collect coarse- and fine-grained information (see \textsection \ref{sec:annotation-process}) on CSW research in NLP that includes languages covered (see \textsection \ref{sec:languages}), NLP tasks that have been explored, and new and emerging trends (see \textsection \ref{sec:tasks}). In addition, motivated by the fact that fields like linguistics, socio-linguistics, and related fields, have studied CSW since the early 1900s, we also investigate to what extent theoretical frameworks from these fields have influenced NLP approaches (see \textsection \ref{sec:linguistics}), and how the choice of methods has evolved over time (see \textsection \ref{sec:methods}). Finally, we discuss the most pressing research challenges and identify a path forward to continue advancing this exciting line of work (see \textsection \ref{sec:futurework}). 

The area of NLP for CSW data is thriving, covering an increasing number of language combinations and tasks, and it is clearly advancing from a niche field to a common research topic, thus making our comprehensive survey timely. We expect the survey to provide valuable information to researchers new to the field and motivate more research from researchers already engaging in NLP for CSW data.  %in an easy to digest manner, regarding the following aspects: 1) past efforts in the field; 2) existing resources and languages covered; 3) recent trends in methodologies; 4) as well as recommendations for future work to continue advancing this exciting line of work. 

% We summarize our major contributions of this work are as following:
% \begin{itemize}
%     \item We conduct a systematic survey and human annotations on more than 350 papers published in open-source and major NLP and speech proceedings to collect coarse-grained and fine-grained information on CSW research in NLP.
%     \item We study the trend and challenges from the CSW research that has been done for decades.
%     \item We provide suggestions and feedback on future directions and rooms for improvement in the CSW research field.
% \end{itemize}

\begin{figure}[!t]
    \centering
    \includegraphics[width=\linewidth]{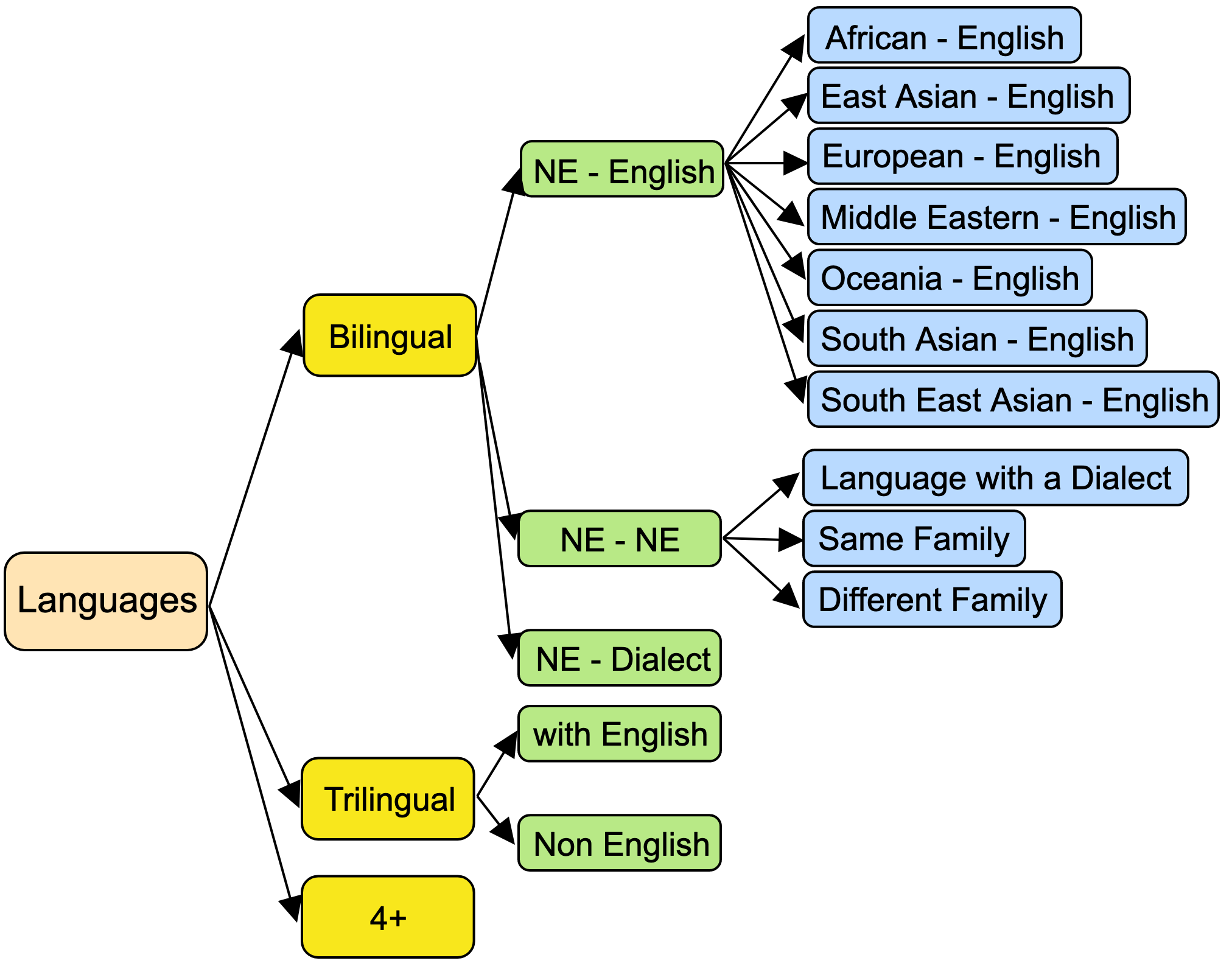}   
	\caption{Language Categories. *NE denotes Non English. We show fine-grained categories in \textbf{green} and \textbf{blue}.}
	\label{fig:languages}
\end{figure}

\section{Exploring Open Proceedings}
\label{sec:annotation}

To develop a holistic understanding of the trends and advances in CSW NLP research, we collect research papers on CSW from the ACL Anthology and ISCA proceedings. We focus on these two sources because they encompass the top venues for publishing in speech and language processing in our field. In addition, we also look into personal repositories from researchers in the community that contains a curated list of CSW-related papers. We discuss below the search process for each venue.

%\subsection{Data Collection}
\paragraph{ACL Anthology}
We crawled the entire the ACL Anthology repository up to October 2022.\footnote{\url{https://github.com/acl-org/acl-anthology}} We then filtered papers by using the following keywords related to CSW:
%\begin{itemize}
 %   \item \textbf{Code-Switching keywords:} 
 ``codeswitch", ``code switch", ``code-switching", ``code-switched", ``code-switch", ``code-mix", ``code-mixed", ``code-mixing", ``code mix"
   % \item \textbf{Mixed Language keywords:} 
   ``mixed-language", ``mixed-lingua", ``mixed language", ``mixed lingua", and ``mix language". 
%\end{itemize}

\paragraph{ISCA Proceedings}
We manually reviewed publicly available proceedings on the ISCA website\footnote{\url{https://www.isca-speech.org}} and searched for papers related to CSW using the same set of keywords as above.

\begin{figure}[!t]
    \centering
    \begin{subfigure}{0.85\linewidth}
		\centering
		\includegraphics[width=\linewidth]{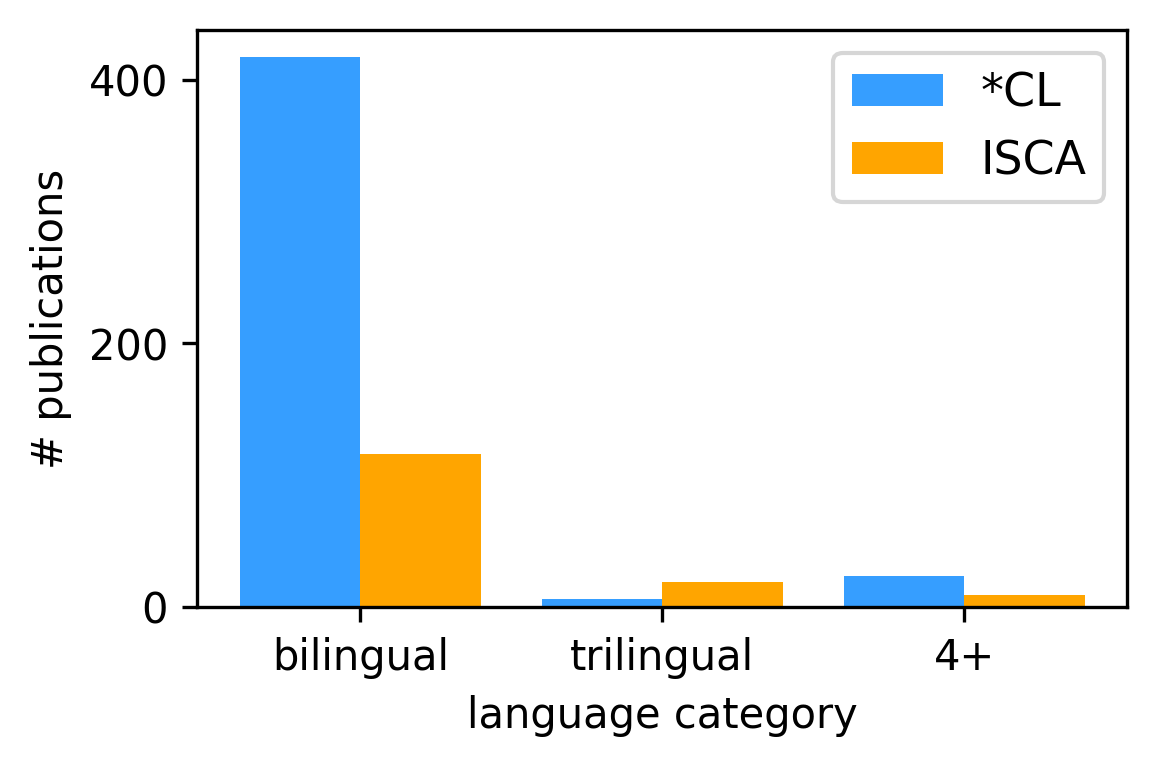}   
	\end{subfigure}
	\begin{subfigure}{0.9\linewidth}
		\centering
        \includegraphics[width=0.95\linewidth]{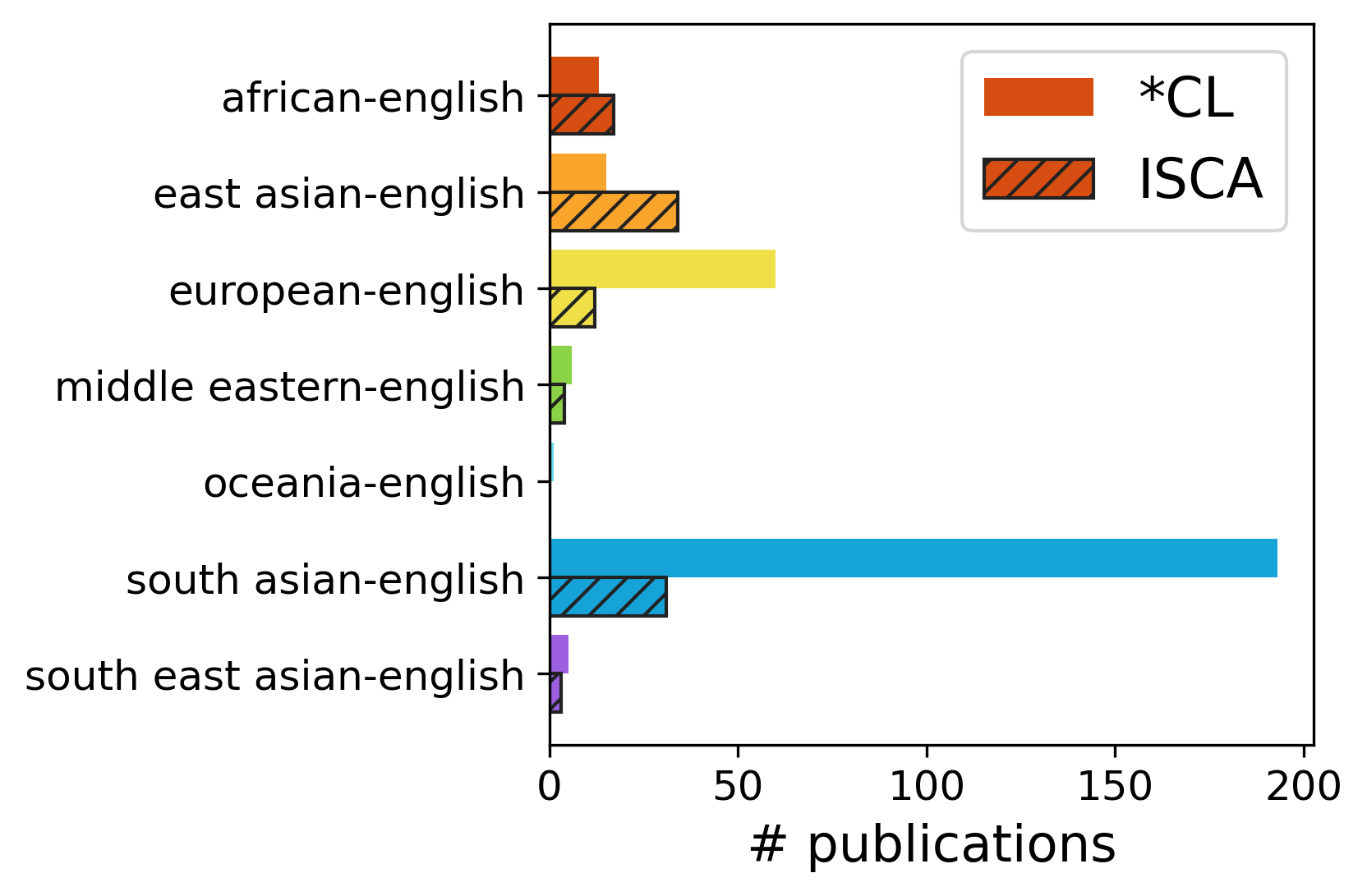} 
	\end{subfigure}
	\caption{\textbf{(Top):} Number of publications across the type of language combination (bilingual, trilingual or 4+. \textbf{(Bottom):} Number of publications on fine-grained bilingual category with English as the L2 language.}
	\label{fig:language_publications}
\end{figure}

\begin{figure}[!t]
    \centering
    \includegraphics[width=\linewidth]{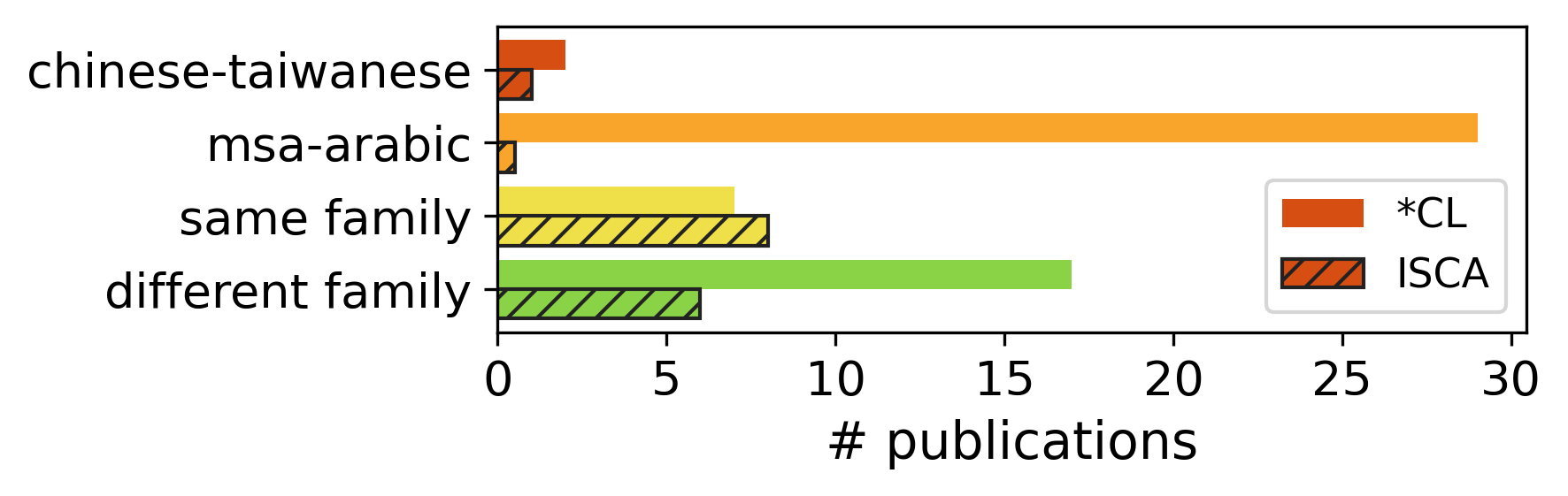}  
	\caption{Number of publications of bilingual code-switched languages that do not contain English. $^*$msa stands for Modern Standard Arabic. The first two are the combination of a language with its dialect.}
	\label{fig:language_publications_non_english}
\end{figure}

\paragraph{Web Resources}
To extend the coverage of paper sources, we also gathered data from existing repositories.\footnote{\url{https://github.com/gentaiscool/code-switching-papers}}\footnote{\url{https://genius1237.github.io/emnlp19_tut/}} We can find multiple linguistics papers studying about CSW.

\subsection{Annotation Process}
\label{sec:annotation-process}
We have three human annotators to annotate all collected papers based on multiple categories shown in Table~\ref{tab:category}. All papers are coded by a least one annotator. To extract the specific information we are looking for from the paper, the annotator needs to read through the paper, as most of the information is not contained in the abstract. The full list of the annotations we collected is available in the Appendix (see \textsection \ref{sec:annotation-catalog}).

To facilitate our analysis, we annotated the following aspects:
\begin{itemize}
    \item \textbf{Languages:} CSW is not restricted to pairs of languages; thus, we divide papers by the number of languages that are covered into \texttt{bilingual}, \texttt{trilingual}, and \texttt{4+} (if there are at least four languages). For a more fine-grained classification of languages, we categorize them by geographical location (see Figure~\ref{fig:languages}). 
    \item \textbf{Venues:} There are multiple venues for CSW-related publications. We considered the following type of venues: conference, workshop, symposium and book. As we will discuss later, the publication venue is a reasonable data point of the wider spread of CSW research in recent years. 
    \item \textbf{Papers:} We classify the paper types based on their contribution and nature. We predict that we will have a high distribution of dataset/resource papers, as lack of resources has been a major bottleneck in the past.
    \item \textbf{Datasets:} If the paper uses a dataset for the research, we will identify the source and modality (i.e., written text or speech) of the dataset. 
    \item \textbf{Methods:} We identify the type of methods presented in work. 
    \item \textbf{Tasks:} We identify the downstream NLP tasks (including the speech processing-related tasks) presented in work.
\end{itemize}

\begin{table}[!t]
  \centering
  \resizebox{0.5\textwidth}{!}{
  \begin{tabular}{lccc|c}
    \toprule
        \multirow{2}{*}{\textbf{Languages}} & \multicolumn{4}{c}{\textbf{\# Publications}} \\ 
        & \textbf{*CL} & \textbf{ISCA} & \textbf{Total} & \textbf{Shared Task} \\ \midrule
        Hindi-English & 111 & 17 & 128 & 30 \\
        Spanish-English & 78 & 8 & 86 & 40 \\
        Chinese-English$^\ddagger$ & 20 & 27 & 47 & 5 \\
        Tamil-English & 37 & 2 & 39 & 17 \\
        Malayalam-English & 23 & 2 & 25 & 13 \\
    \bottomrule
  \end{tabular}
  }
  \caption{Most common code-switching languages in *CL and ISCA venues. $^\ddagger$The count does not include the dialect or South East Asian Mandarin - English~\cite{lyu2010seame} since they contain more than two languages (i.e., it has words from the Chinese dialect).} 
  \label{top-languages}
\end{table}

\section{Language Diversity}
\label{sec:languages}

Here, we show the languages covered in the CSW resources. While focusing on the CSW phenomenon increases diversity of NLP technology, as we will see in this section, future efforts are needed to provide significant coverage of the most common CSW language combinations worldwide.

\begin{table}[!t]
  \centering
  \resizebox{0.49\textwidth}{!}{
  \begin{tabular}{lccc}
    \toprule
        \textbf{Languages} & \multicolumn{3}{c}{\textbf{\# Publications}} \\ 
        & \textbf{non-ST} & \textbf{ST} & \textbf{Total} \\
        \midrule
        Language Identification & 46 & 17 & 63 \\
        Sentiment Analysis & 31 & 30 & 61 \\
        NER & 17 & 14 & 31\\
        POS Tagging & 29 & 1 & 30\\
        Abusive/Offensive Lang. Detection & 9 & 16 & 25 \\
        ASR & 20 & 0 & 22 \\
        Language Modeling & 19 & 1 & 20 \\
        Machine Translation & 8 & 5 & 13 \\
    \bottomrule
  \end{tabular}
  }
  \caption{Most common tasks in ACL venues. ST denotes shared task.} 
  \label{top-tasks}
\end{table}

\subsection{Variety of Mixed Languages}
Figure~\ref{fig:language_publications} shows the distribution of languages represented in the NLP for CSW literature. Most of the papers use datasets with two language pairs. However, we did find a few papers that address CSW scenarios with more than two languages. We consider this a relevant future direction in CSW: scaling model abilities to cover n languages, with $n \ge 2$. 

\paragraph{CSW in two languages} We group the number of publications focusing on bilingual CSW based on world regions in Figure~\ref{fig:language_publications} (bottom). We can see that the majority of research in CSW has focused on South Asian-English, especially on Hindi-English, Tamil-English, and Malayalam-English, as shown in Table~\ref{top-languages}. The other common language pairs are Spanish-English and Chinese-English. That table also shows that many of the publications are shared task papers. This is probably reflecting efforts from a few research groups to motivate more research into CSW, such as that behind the CALCS workshop series.

Looking at the languages covered, we also find that there are many language pairs that come from different language families, such as Turkish-German~\cite{ccetinouglu2016turkish,ccetinouglu2019challenges,ozatecs2021language,ozatecs2022improving}, Turkish-Dutch~\cite{gamback2016comparing}, French-Arabic~\cite{sankoff1998production,lounnas2021towards}, Russian-Tatar~\cite{taguchi2021transliteration}, Russian-Kazakh~\cite{mussakhojayeva2022kazakhtts2}, Hindi-Tamil~\cite{thomas2018code}, Arabic-North African~\cite{el2018arabic}, Basque-Spanish~\cite{aguirre2022basco}, and Wixarika-Spanish~\cite{mager2019subword}. There are only very few papers working on Middle Eastern - English language pairs, most of the time, the Middle Eastern languages are mixed with non-English and/or dialects of these languages (see Figure~\ref{fig:language_publications_non_english}).

\paragraph{Trilingual} The number of papers addressing CSW in more than two languages is still small (see \ref{fig:language_publications} top), compared to the papers looking at pairs of languages. Not surprisingly, this smaller number of papers focus on world regions where either the official languages are more than two, or these languages are widely used in the region, for example, Arabic-English-French~\cite{abdul2020toward}, Hindi-Bengali-English~\cite{barman2016part}, Tulu-Kannada-English~\cite{hegde2022corpus}, and Darija-English-French~\cite{voss2014finding}.

\paragraph{4+} When looking at the papers that  focus on more than three languages, we found that many papers use South East Asian Mandarin-English (SEAME) dataset~\cite{lyu2010seame}, which has Chinese dialects and Malay or Indonesian words. Most of the other datasets are machine-generated using rule-based or neural methods.

\subsection{Language-Dialect Code-Switching}
Based on Figure~\ref{fig:language_publications_non_english}, we can find some papers with language-dialect CSW, such as Chinese-Taiwanese Dialect~\cite{chu2007language,yu2012language} and Modern Standard Arabic (MSA)-Arabic Dialect~\cite{elfardy2012token,samih2016arabic,el2018arabic}. The dialect, in this case, is the variation of the language with a different form that is very specific to the region where the CSW style is spoken.

\section{Tasks and Datasets}
\label{sec:tasks}
In this section, we summarize our findings, focusing on the CSW tasks and datasets. Table~\ref{top-tasks} shows the distribution of CSW tasks for ACL papers with at least ten publications. The two most popular tasks are language identification and sentiment analysis. Researchers mostly use the shared tasks from 2014~\cite{solorio2014overview} and 2016~\cite{molina2016overview} for language identification, and the SemEval 2020 shared task~\cite{patwa2020semeval} for sentiment analysis. For ISCA, the most popular tasks are unsurprisingly ASR and TTS. This strong correlation between task and venue shows that the speech processing and *CL communities remain somehow fragmented and working in isolation from one another, from the most part.

 % In general, the shared tasks on language identification from 2014~\cite{solorio2014overview} and 2016~\cite{molina2016overview} contribute to the ACL publications. 

\begin{table}[!t]
    \centering
    \resizebox{0.4\textwidth}{!}{
    \begin{tabular}{lrrr}
    \toprule
      & \multicolumn{3}{c}{\textbf{\# Publications}} \\
         & \textbf{*CL} & \textbf{ISCA} & \textbf{Total} \\
         \midrule
         Public Dataset & 38 & 4 & 42\\
         Private Dataset & 54 & 18 & 72 \\
         \bottomrule
    \end{tabular}
    }
    \caption{Publications that introduce new corpus.}
    \label{tab:corpus}
\end{table}

\begin{table}[!t]
    \centering
    \resizebox{0.49\textwidth}{!}{
    \begin{tabular}{lccc}
    \toprule
         \textbf{Source} & \textbf{*CL} & \textbf{ISCA} & \textbf{Total} \\
         \midrule
         Social Media & 183 & 3 & 186
         \\ 
         Speech (Recording) & 29 & 102 & 141 \\
         Transcription & 23 & 4 & 27 
         \\ 
         News & 19 & 5 & 24 \\ 
         Dialogue & 16 & 2 & 18 \\
         Books & 7 & 1 & 8 \\
         Government Document & 6 & 0 & 6 \\
         Treebank & 5 & 0 & 5 \\
         \bottomrule
    \end{tabular}
    }
    \caption{The source of the CSW dataset in the literature.}
    \label{tab:datasets-source}
\end{table}

\paragraph{Public vs. Private Datasets}
Public datasets availability also dictates what tasks are being explored in CSW research. Public datasets such as HinGE~\cite{srivastava2021hinge}, SEAME~\cite{lyu2010seame} and shared task datasets~\cite{solorio2014overview,molina2016overview,aguilar2018named,patwa2020semeval} have been widely used in many of the papers. Some work, however, used new datasets that are not publicly available, thus hindering adoption (see Table~\ref{tab:corpus}). There are two well-known benchmarks in CSW: LinCE~\cite{aguilar2020lince} and GlueCOS~\cite{khanuja2020gluecos}. These two benchmarks have a handful of tasks, and they are built to encourage transparency and reliability of evaluation since the test set labels are not publicly released. The evaluation is done automatically on their websites. However, their support languages are mostly limited to popular CSW language pairs, such as Spanish-English, Modern Standard Arabic-Egyptian, and Hindi-English, the exception being Nepali-English in LinCE.

\paragraph{Dataset Source}
Table~\ref{tab:datasets-source} shows the statistics of dataset sources in the CSW literature. We found that most of the ACL papers were working on social media data. This is expected, considering that social media platforms are known to host informal interactions among users, making them reasonable places for users to code-switch. Naturally, most ISCA papers work on speech data, many of which are recordings of conversations and interviews. There are some datasets that come from speech transcription, news, dialogues, books, government documents, and treebanks.

\paragraph{Paper Category}
Table~\ref{tab:paper-type} presents the distribution of CSW papers. Most of the papers are empirical work focusing on the evaluation of downstream tasks. The second largest population is shared tasks. We also notice that many papers introduce new CSW corpus, but they are not released publicly. Some papers only release the URL or id to download the datasets, especially for datasets that come from social media (e.g., Twitter) since redistribution of the actual tweets is not allowed~\cite{solorio2014overview,molina2016overview} resulting in making reproducibility harder. Social media users can delete their posts at any point in time, resulting in considerable data attrition rates. There are very few papers working on the demos, theoretical work, position papers, and introducing evaluation metrics. 

\section{From Linguistics to NLP}
\label{sec:linguistics}
Notably, papers are working on approaches that are inspired by linguistic theories to enhance the processing of CSW text. In this survey, we find three linguistic constraints that are used in the literature: equivalence constraint, matrix-embedded language Framework (MLF), and Functional Head Constraint. In this section, we will briefly introduce the constraints and list the papers that utilize the constraints.

\begin{table}[!t]
    \centering
    \resizebox{0.49\textwidth}{!}{
    \begin{tabular}{lccc}
    \toprule
         \textbf{Type} & \textbf{*CL} & \textbf{ISCA} & \textbf{Total} \\
         \midrule
         Empirical & 205 & 100 & 305 \\
         Shared Task & 82 & 1 & 83 \\
         Corpus (Closed) & 54 & 18 & 62 \\
         Corpus (Open) & 38 & 4 & 42 \\
         Analysis & 34 & 8 & 42 \\
         Demo & 7 & 2 & 9 \\
         Theoretical/Linguistic & 7 & 0 & 7 \\
         Position/Opinion/Survey & 3 & 0 & 3 \\
         Metric & 2 & 1 & 3 \\
         \bottomrule
    \end{tabular}
    }
    \caption{Paper Type of the CSW papers.} 
    % One paper can be attributed to more than one type.}
    \label{tab:paper-type}
\end{table}

\subsection{Linguistic-Driven Approaches}

\paragraph{Equivalence Constraint} In a well-formed code-switched sentence, the switching takes place at those points where the grammatical constraints of both languages are satisfied \cite{poplack1980sometimes}. \citet{li2012code,li2013language} incorporate this syntactic constraint to a statistical code-switch language model (LM) and evaluate the model on Chinese-English code-switched speech recognition. On the same line of work, \citet{pratapa2018language,pratapa2021comparing} implement the same constraint to Hindi-English CSW data by producing parse trees of parallel sentences and matching the surface order of child nodes in the trees. \citet{winata2019code} apply the constraint to generate synthetic CSW text and find that combining the real CSW data with synthetic CSW data can effectively improve the perplexity. They also treat parallel sentences as a linear structure and only allow switching on non-crossing alignments.

% This constraint usually holds for syntactically similar languages such as English-Spanish but not for syntactically divergent languages such as Hebrew-Spanish \cite{berk1986linguistic}.

\paragraph{Matrix-Embedded Language Framework (MLF)}
\citet{myers1997duelling} proposed that in bilingual CSW, there exists an asymmetrical relationship between the dominant \textit{matrix language} and the subordinate \textit{embedded language}. Matrix language provides the frame of the sentence by governing all or most of the most of the grammatical morphemes as well as word order, whereas syntactic elements that bear no or only limited grammatical function can be provided by the embedded language \cite{johanson1999dynamics,myers2005multiple}. % Closed-class items, such as prepositions and auxiliary verbs, cannot be replaced by their
% equivalents in the embedded language \cite{joshi1982processing}. 
\citet{lee2019linguistically} use augmented parallel data by utilizing MLF to supplement the real code-switch data. \citet{gupta2020semi} use MLF to automatically generate the code-mixed text from English to multiple languages without any parallel data. 

\paragraph{Functional Head Constraint} \citet{belazi1994code} posit that it is impossible to switch languages between a functional head and its complement because of the strong relationship between the two constituents. ~\citet{li2014language} use the constraint of the LM by first expanding the search network with a translation model and then using parsing to restrict paths to those permissible under the constraint.

\subsection{Learning from Data Distribution}

Linguistic constraint theories have been used for decades to generate synthetic CSW sentences to address the lack of data issue. However, the approach requires external word alignments or constituency parsers that create erroneous results instead of applying the linguistic constraints to generate new synthetic CSW data, building a pointer-generator model to learn the real distribution of code-switched data~\cite{winata2019code}. \citet{chang2019code} propose to generate CSW sentences from monolingual sentences using Generative Adversarial Network (GAN)~\cite{goodfellow2020generative} and the generator learns to predict CSW points without any linguistic knowledge.

\subsection{The Era of Statistical Methods}
The research on CSW is also influenced by the progress and development of machine learning. According to Figure~\ref{fig:methods}, starting in 2006, statistical methods have been adapted to CSW research, while before that year, the approaches were mainly rule-based. There are common statistical methods for text classification used in the literature, such as Naive Bayes~\cite{solorio2008learning} and Support Vector Machine (SVM)~\cite{solorio2008part}. Conditional Random Field (CRF)~\cite{sutton2012introduction} is also widely seen in the literature for sequence labeling, such as Part-of-Speech (POS) tagging~\cite{vyas2014pos}, Named Entity Recognition (NER), and word-level language identification~\cite{lin2014cmu,chittaranjan2014word,jain2014language}. HMM-based models have been used in speech-related tasks, such as speech recognition~\cite{weiner2012integration,li2013language} and text synthesis~\cite{qian2008hmm,shuang2010hmm,he2012turning}.

\subsection{Utilizing Neural Networks}
\label{sec:methods}

\begin{figure}[!t]
    \centering
    \begin{subfigure}{0.95\linewidth}
		\centering
		\includegraphics[width=\linewidth]{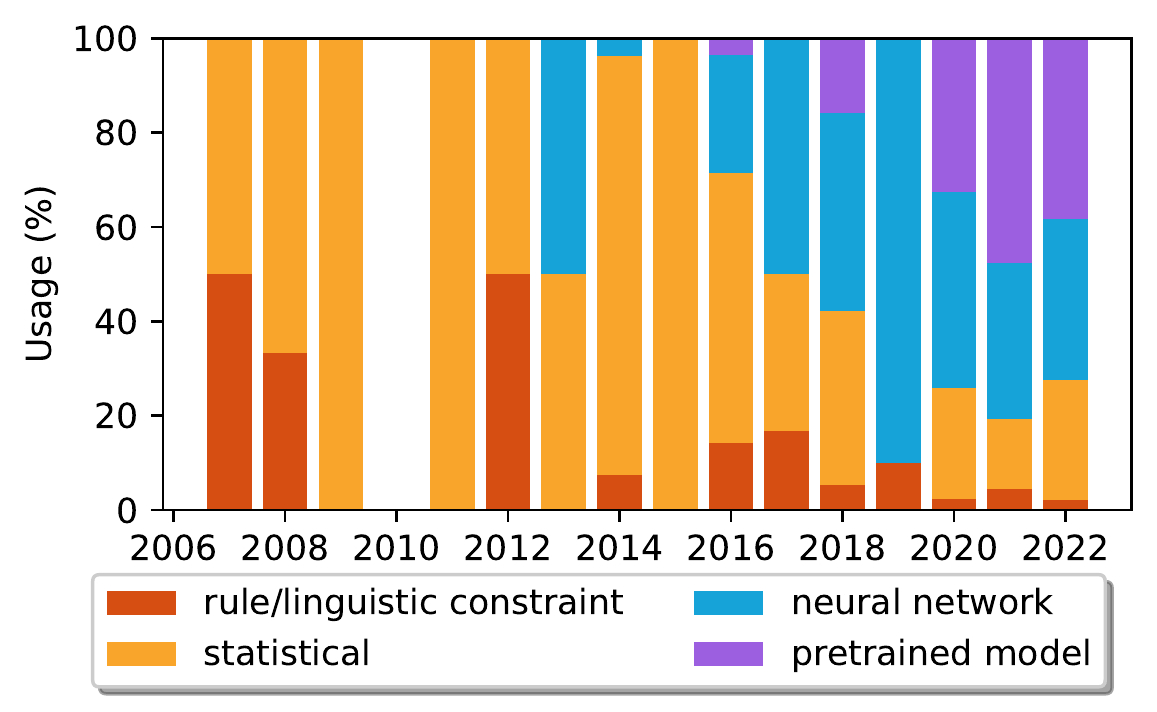}   
		\caption{*CL}
	\end{subfigure}
	\begin{subfigure}{0.95\linewidth}
		\centering
        \includegraphics[width=\linewidth]{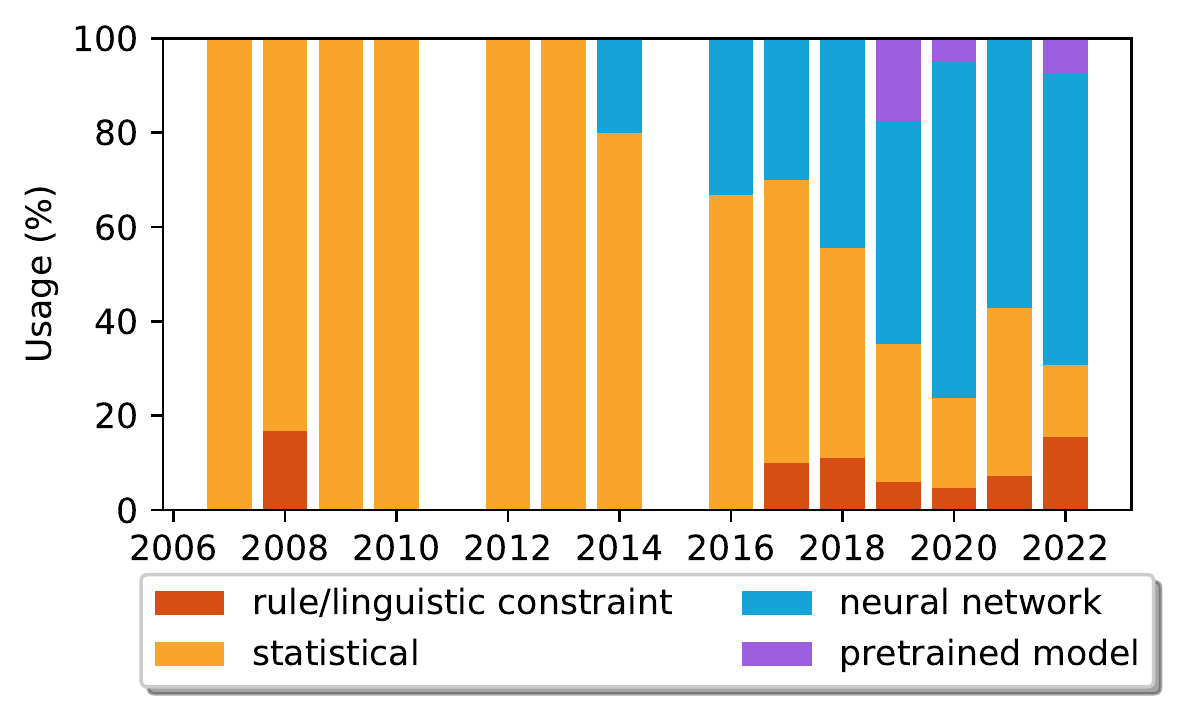} 
        \caption{ISCA}
	\end{subfigure}
	\caption{Methods used for code-mixing NLP.} % over the years
	\label{fig:methods}
\end{figure}

Following general NLP trends, we see the adoption of neural methods and pre-trained models growing in popularity over time. In contrast, the statistical and rule-based approaches are diminishing. Compared to ISCA, we see more adaptation of the pre-training model. This is because ACL work is more text-based focused, where pre-trained LMs are more widely available.

\paragraph{Neural-Based Models}
Figure~\ref{fig:methods} shows that the trend of using neural-based models started in 2013, and the usage of rule/linguistic constraint and statistical methods diminished gradually through time, but they are still used even with a low percentage. RNN and LSTM architectures are commonly used in sequence modeling, such as language modeling~\cite{adel2013combination,vu2014exploration,adel2014combining,winata2018code,garg2018code,winata2019code} and CSW identification~\cite{samih2016multilingual}. DNN-based and hybrid HMM-DNN models are used in speech recognition models~\cite{yilmaz2018acoustic,yilmaz2018building}.

\paragraph{Pre-trained Embeddings} 

Pre-trained embeddings are used to complement neural-based approaches by initializing the embedding layer.
Common pre-trained embeddings used in the literature are monolingual subword-based embeddings, FastText~\cite{joulin2016fasttext}, and aligned-embeddings MUSE~\cite{conneau2017word}. A standard method to utilize monolingual embeddings is to concatenate or sum two or more embeddings from different languages~\cite{trivedi2018iit}. A more recent approach is to apply an attention mechanism to merge embeddings and form meta-embeddings~\cite{winata2019learning,winata2019hierarchical}. Character-based embeddings have also been explored in the literature to address the out-of-vocabulary issues on word-embeddings~\cite{winata2018bilingual,attia2018ghht,aguilar2021char2subword}. Another approach is to train bilingual embeddings using real and synthetic CSW data~\cite{pratapa2018word}. In the speech domain, ~\citet{lovenia2022ascend} utilize wav2vec 2.0~\cite{baevski2020wav2vec} as a starting model before fine-tuning.

\paragraph{Language Models}

Many pre-trained model approaches utilize multilingual LMs, such as mBERT or XLM-R to deal with CSW data~\cite{khanuja2020gluecos,aguilar2020english,pant2020towards,patwa2020semeval,winata2021multilingual}. 
% Some use language-specific language models. 
These models are often fine-tuned with the downstream task or with CSW text to better adapt to the languages. Some downstream fine-tuning approaches use synthetic CSW data due to a lack of available datasets. ~\citet{aguilar2021char2subword} propose a character-based subword module (char2subword) of the mBERT that learns the subword embedding that is suitable for modeling the noisy CSW text. \citet{winata2021multilingual} compare the performance of the multilingual LM versus the language-specific LM for CSW context. While XLM-R provides the best result, it is also computationally heavy. There needed to be more exploration of larger models. 

We see that pre-trained LMs provide better empirical results on current benchmark tasks and enables an end-to-end approach. Therefore, one can theoretically work on CSW tasks without any linguistic understanding of the language, assuming the dataset for model finetuning is available. However, the downside is that there is little understanding of how and when the LMs would fail, thus we encourage more interpretability work on these LMs in CSW setting.

\section{Recent Challenges and Future Direction}
\label{sec:futurework}
\subsection{More Diverse Exploration on Code-Switching Styles and Languages}

A handful of languages, such as Spanish-English, Hindi-English, or Chinese-English, dominate research and resource CSW. However, there are still many countries and cultures rich in the use of CSW, which is still under-represented in NLP research~\cite{joshi2020state,aji2022one,yong2023prompting}, especially on different CSW variations. CSW style can vary in different regions of the world, and it would be interesting to gather more datasets on unexplored and unknown styles, which can be useful for further research and investigation on linguistics and NLP.
% Some of them even have a massive number of speakers. 
Therefore, one future direction is to broaden the language scope of CSW research.

\subsection{Datasets: Access and Sources}
According to our findings, there are more than 60\% of the datasets are private (see Table~\ref{tab:corpus}), and they are not released to the public. This eventually hampers the progress of CSW research, particularly in the results' reproducibility, credibility, and transparency. Moreover, many studies in the literature do not release the code to reproduce their work. Therefore, we encourage researchers who build a new corpus to release the datasets publicly. In addition, the fact that some researchers provide urls to download the data is also problematic due to the data attrition issue we raised earlier. Data attrition is bad for reproducibility, but it is also a waste of annotation efforts. Perhaps we should work on identifying alternative means to collect written CSW data in an ecologically valid manner.

\subsection{Model Scaling}
To the best of our knowledge, little work has been done on investigating how well the scaling law holds for code-mixed datasets. \citet{winata2021multilingual} demonstrate that the XLM-R-large model outperforms smaller pre-trained models on the NER and POS tasks in LinCE benchmark~\cite{aguilar2020lince}; however, the largest model in the study, which is the XLM-R-large model, only has 355 million parameters. Furthermore, they find that smaller models that combine word, subword, and character embeddings achieve comparable performance as mBERT while being faster in inference. Given the recent release of billion-sized large pre-trained multilingual models such as XGLM and BLOOM \cite{scao2022bloom}, we urge future research to study the scaling law and performance-compute trade-off in code-mixing tasks. 

\subsection{Zero-Shot and Few-Shot Exploration}
The majority of pre-trained model approaches fine-tune their models to the downstream task. On the other hand, CSW data is considerably limited. With the rise of multilingual LMs, especially those that have been fine-tuned with prompt/instruction~\cite{muennighoff2022crosslingual,ouyang2022training,winata2022cross}, one direction is to see whether these LMs can handle CSW input in a zero-shot fashion. This work might also tie in with model scaling since larger models have shown better capability at zero-shot and few-shot settings~\cite{winata2021language,srivastava2022beyond}.

\subsection{Robustness Evaluation}
Since CSW is a widely common linguistic phenomenon, we argue that cross-lingual NLP benchmarks, such as XGLUE \cite{liang2020xglue} and XTREME-R \cite{ruder2021xtreme}, should incorporate linguistic CSW evaluation \cite{aguilar2020lince,khanuja2020gluecos}. The reasons are that CSW is a cognitive ability that multilingual human speakers can perform with ease \cite{beatty2020codeswitching}. CSW evaluation examines the robustness of multilingual LMs in learning cross-lingual alignment of representations \cite{conneau2020emerging,libovicky2020language,pires2019multilingual,adilazuarda2022indorobusta}. On the other hand, catastrophic forgetting is observed in pre-trained models \cite{shah2020learning}, and human speakers \cite[known as language attrition]{hirvonen2000code,du2009language} in a CSW environment. We argue that finetuning LMs on code-mixed data is a form of \textit{continual learning} to produce a more generalized multilingual LM. Thus, we encourage CSW research to report the performance of finetuned models on both CSW and monolingual texts. 

\subsection{Task Diversity}
We encourage creating reasoning-based tasks for CSW texts for two reasons. First, code-mixed datasets for tasks such as NLI, coreference resolution, and question-answering are much fewer in comparison to tasks such as sentiment analysis, parts-of-speech tagging, and named-entity recognition. Second, comprehension tasks with the CSW text present more processing costs for human readers \cite{bosma2020switching}.

\subsection{Conversational Agents}
There has been a recent focus on developing conversational agents with LMs such as ChatGPT\footnote{\url{https://openai.com/blog/chatgpt/}}, Whisper~\cite{radford2022robust}, SLAM~\cite{bapna2021slam}, mSLAM~\cite{bapna2022mslam}.
% (and its multilingual version, mSLAM~\cite{bapna2022mslam}). 
We recommend incorporating the capability of synthesizing code-mixed data in human-machine dialogue, as CS is a prevalent communication style among multilingual speakers \cite{ahn2020code}, and humans prefer chatbots with such capability \cite{bawa2020multilingual}. 

\subsection{Automatic Evaluation for Generation}
With the rise of pre-trained models, generative tasks gained more popularity. However, when generating CSW data, most work used human evaluation for measuring quality of the generated data. Alternative automatic methods for CS text based on word-frequency and temporal distribution are commonly used \cite{guzman2017metrics,mave2018language}, but we believe there is still much room for improvement in this respect. One possible future direction is to align the evaluation metrics to human judgement of quality \cite{hamed2022benchmarking} where we can assess separately the ``faithfulness'' of the resulting CSW data from other desired properties of language generation. Other nuances here are related to the intricacy of CSW patterns, where ideally the model would mimic the CSW style of the intended users.

\section{Conclusion}
We present a comprehensive systematic survey on code-switching research in natural language processing to explore the progress of the past decades and understand the existing challenges and tasks in the literature. We summarize the trends and findings and conclude with a discussion for future direction and open questions for further investigation. We hope this survey can encourage and lead NLP researchers in a better direction on code-switching research.

\section*{Limitations}

The numbers in this survey are limited to papers published in the ACL Anthology and ISCA Proceedings. However, we also included papers as related work from other resources if they are publicly available and accessible. In addition, the category in the survey does not include the code-switching type (i.e., intra-sentential, inter-sentential, etc.) since some papers do not provide such information. 

\section*{Ethics Statement}
We use publicly available data in our survey with permissive licenses. No potential ethical issues in this work. 

\section*{Acknowledgements}

Thanks to Igor Malioutov for the insightful discussion on the paper.

% A

% Entries for the entire Anthology, followed by custom entries
% \bibliography{anthology-2,custom}
\bibliography{custom}
\bibliographystyle{acl_natbib}

\appendix

\section{Annotation Catalog} 
\label{sec:annotation-catalog} 
We release the annotation of all papers we use in the survey.

\subsection{*CL Anthology}
\paragraph{Bilingual}
Table~\ref{tab:acl-catalog-african-english} shows the annotation for papers with African-English. Table~\ref{tab:acl-catalog-east-asian-english} shows the annotation for papers with East Asian-English languages. Table~\ref{tab:acl-catalog-european-english} shows the annotation for papers with European-English languages. Table~\ref{tab:acl-catalog-mideast-english} shows the annotation for papers with Middle Eastern-English languages. Table~\ref{tab:acl-catalog-south-asian-english-1} and Table~\ref{tab:acl-catalog-south-asian-english-2} show the annotation for papers with South Asian-English languages. Table~\ref{tab:acl-catalog-sea-english} shows the annotation for papers with South East Asian-English languages. Table~\ref{tab:acl-catalog-dialect} shows the annotation for papers with a combination of a language with a dialect. Table~\ref{tab:acl-catalog-samefam} shows the annotation for papers with languages in the same family. Table~\ref{tab:acl-catalog-diffam} shows the annotation for papers with languages in different families.

\paragraph{Trilingual}
Table~\ref{tab:acl-catalog-3lang} shows the annotation for papers with three languages.

\paragraph{4+}
Table~\ref{tab:acl-catalog-4lang} shows the annotation for papers with four or more languages.

\subsection{ISCA Proceeding}
\paragraph{Bilingual}
Table~\ref{tab:isca-catalog-african-english} shows the annotation for papers with African-English. Table~\ref{tab:isca-catalog-eastasian-english} shows the annotation for papers with East Asian-English languages. Table~\ref{tab:isca-catalog-european-english} shows the annotation for papers with European-English languages. Table~\ref{tab:isca-catalog-middleeastern-english} shows the annotation for papers with Middle Eastern-English languages. Table~\ref{tab:isca-catalog-southasian-english} shows the annotation for papers with South Asian-English languages. Table~\ref{tab:isca-catalog-southeastasian-english} shows the annotation for papers with South East Asian-English languages. Table~\ref{tab:isca-catalog-language-with-dialects} shows the annotation for papers with a combination of a language with a dialect. Table~\ref{tab:isca-catalog-two-langauges-in-the-same-family} shows the annotation for papers with languages in the same family. Table~\ref{tab:isca-catalog-two-langauges-in-different-families} shows the annotation for papers with languages in different families.

\paragraph{Trilingual}
Table~\ref{tab:isca-catalog-trilingual} shows the annotation for papers with three languages.

\paragraph{4+}
Table~\ref{tab:isca-catalog-4+} shows the annotation for papers with four or more languages.

\begin{table*}[!t]
    \centering
    \resizebox{\linewidth}{!}{
\begin{tabular}{@{}lcccccc@{}} \toprule
        	\textbf{Paper} & \textbf{Proceeding}		&	\textbf{IsiZulu}	&	\textbf{Swahili}	&	\textbf{isiXhosa}	&	\textbf{Setswana}	&	\textbf{Sesotho}	\\
         \midrule
	&		&	5	&	1	&	3	&	3	&	3	\\
 \midrule
\cite{joshi1982processing} &	COLING	&	\cellcolor{blue!25}$\checkmark$	&		&		&		&		\\
\cite{piergallini2016word} &	CALCS	&		&	\cellcolor{blue!25}$\checkmark$	&		&		&		\\
\cite{niesler2018first} &	LREC	&	\cellcolor{blue!25}$\checkmark$	&		&	\cellcolor{blue!25}$\checkmark$	&	\cellcolor{blue!25}$\checkmark$	&	\cellcolor{blue!25}$\checkmark$	\\
\cite{biswas2020semi} &	CALCS	&	\cellcolor{blue!25}$\checkmark$	&		&		&		&		\\
\cite{wilkinson2020semi} &	SLTU and CCURL	&	\cellcolor{blue!25}$\checkmark$	&		&	\cellcolor{blue!25}$\checkmark$	&	\cellcolor{blue!25}$\checkmark$	&	\cellcolor{blue!25}$\checkmark$	\\
\cite{biswas2020semi} &	LREC	&	\cellcolor{blue!25}$\checkmark$	&		&	\cellcolor{blue!25}$\checkmark$	&	\cellcolor{blue!25}$\checkmark$	&	\cellcolor{blue!25}$\checkmark$	\\ \bottomrule
        \end{tabular}
    }
    \caption{*CL Catalog in African-English.}
    \label{tab:acl-catalog-african-english}
\end{table*}

\begin{table*}[!t]
    \centering
    \resizebox{0.85\linewidth}{!}{
        \begin{tabular}{@{}lcccc@{}} \toprule
	\textbf{Paper} &	\textbf{Proceeding}	&	\textbf{Chinese}	&	\textbf{Cantonese}	&	\textbf{Korean}	\\
 \midrule
	&		&	20	&	1	&	1	\\
 \midrule
\cite{fung1999mixed} &	ACL	&	\cellcolor{blue!25}$\checkmark$	&		&		\\
\cite{chan2009automatic}	&	IJCLCLP	&		&	\cellcolor{blue!25}$\checkmark$	&		\\
\cite{li2012mandarin}	&	LREC	&	\cellcolor{blue!25}$\checkmark$	&		&		\\
\cite{peng2014learning} &	ACL-IJCNLP	&	\cellcolor{blue!25}$\checkmark$	&		&		\\
\cite{li2014language} &	EMNLP	&	\cellcolor{blue!25}$\checkmark$	&		&		\\
\cite{solorio2014overview} &	CALCS	&	\cellcolor{blue!25}$\checkmark$	&		&		\\
\cite{chittaranjan2014word} &	CALCS	&	\cellcolor{blue!25}$\checkmark$	&		&		\\
\cite{lin2014cmu} &	CALCS	&	\cellcolor{blue!25}$\checkmark$	&		&		\\
\cite{jain2014language} &	CALCS	&	\cellcolor{blue!25}$\checkmark$	&		&		\\
\cite{king2014iucl+} &	CALCS	&	\cellcolor{blue!25}$\checkmark$	&		&		\\

\cite{huang2014improving} &	EACL	&	\cellcolor{blue!25}$\checkmark$	&		&		\\
\cite{wang2015emotion}	&	ACL-IJCNLP	&	\cellcolor{blue!25}$\checkmark$	&		&		\\
\cite{gamback2016comparing} &	LREC	&	\cellcolor{blue!25}$\checkmark$	&		&		\\
\cite{wang2016bilingual}	&	COLING	&	\cellcolor{blue!25}$\checkmark$	&		&		\\
\cite{ccetinouglu2016challenges} &	CALCS	&	\cellcolor{blue!25}$\checkmark$	&		&		\\
\cite{xia2016accurate} &	CALCS	&	\cellcolor{blue!25}$\checkmark$	&		&		\\
\cite{yang2020csp} &	EMNLP	&	\cellcolor{blue!25}$\checkmark$	&		&		\\
\cite{calvillo2020surprisal} &	EMNLP	&	\cellcolor{blue!25}$\checkmark$	&		&		\\
\cite{lin2020exploring} &	ROCLING	&	\cellcolor{blue!25}$\checkmark$	&		&		\\
\cite{cho2020towards} &	CALCS	&		&		&	\cellcolor{blue!25}$\checkmark$	\\
\cite{lin2021exploiting} &	ROCLING	&	\cellcolor{blue!25}$\checkmark$	&		&		\\
\cite{lovenia2021ascend} &	LREC	&	\cellcolor{blue!25}$\checkmark$	&		&		\\ 

\bottomrule
        \end{tabular}
    }
    \caption{*CL Catalog in East Asian-English.}
    \label{tab:acl-catalog-east-asian-english}
\end{table*}

\begin{table*}[!th]
    \centering
    \resizebox{0.72\linewidth}{!}{
        \begin{tabular}{@{}lcccccccc@{}} \toprule
        	\textbf{Paper} & \textbf{Proceeding} &	\textbf{Spanish}	&	\textbf{French}	&	\textbf{Portugese}	&	\textbf{Polish}	&	\textbf{German}	&	\textbf{Dutch}	&	\textbf{Finnish}	\\ \midrule								
	&		&	78	&	7	&	1	&	1	&	5	&	2	&	1	\\	\midrule					
\cite{sankoff1998production} &	COLING	&		&		&		&		&		&		&	\cellcolor{blue!25}$\checkmark$	\\
\cite{solorio2008learning} &	EMNLP	&	\cellcolor{blue!25}$\checkmark$	&		&		&		&		&		&		\\
\cite{solorio2008part} &	EMNLP	&	\cellcolor{blue!25}$\checkmark$	&		&		&		&		&		&		\\
\cite{peng2014learning}	&	ACL-IJCNLP	&	\cellcolor{blue!25}$\checkmark$	&		&		&		&		&		&		\\
\cite{solorio2014overview}	&	CALCS	&	\cellcolor{blue!25}$\checkmark$	&		&		&		&		&		&		\\
\cite{chittaranjan2014word} &	CALCS	&	\cellcolor{blue!25}$\checkmark$	&		&		&		&		&		&		\\
\bottomrule
        \end{tabular}
    }
    \caption{*CL Catalog in European-English.}
    \label{tab:acl-catalog-european-english}
\end{table*}

\begin{table*}[!t]
    \centering
    \resizebox{0.9\linewidth}{!}{
        \begin{tabular}{@{}lcccccc@{}} \toprule
	\textbf{Paper} & \textbf{Proceeding} &	\textbf{Egyptian Arabic}	&	\textbf{Arabic}	&	\textbf{Turkish}	\\ \midrule
	&		&	3	&	1	&	2	\\ \midrule
\cite{rijhwani2017estimating} &	ACL	&		&		&	\cellcolor{blue!25}$\checkmark$	\\
\cite{hamed2018collection} &	LREC	&	\cellcolor{blue!25}$\checkmark$	&		&		\\
\cite{yirmibecsouglu2018detecting} &	W-NUT	&		&		&	\cellcolor{blue!25}$\checkmark$	\\
\cite{sabty2020contextual} &	WANLP	&		&	\cellcolor{blue!25}$\checkmark$	&		\\
\cite{balabel2020cairo} &	LREC	&	\cellcolor{blue!25}$\checkmark$	&		&		\\
\cite{hamed2020arzen} &	LREC	&	\cellcolor{blue!25}$\checkmark$	&		&		\\
\bottomrule
        \end{tabular}
    }
    \caption{*CL Catalog in Middle Eastern-English.}
    \label{tab:acl-catalog-mideast-english}
\end{table*}

\begin{table*}[!t]
    \centering
    \resizebox{\linewidth}{!}{
        \begin{tabular}{@{}lcccccccccccccccccc@{}} \toprule
	\textbf{Paper} &	 \textbf{Proceeding} &	\textbf{Hindi}	&	\textbf{Marathi}	&	\textbf{Konkani}	&	\textbf{Bengali}	&	\textbf{Bengali}	&	\textbf{Nepali}	&	\textbf{Telugu}	&	\textbf{Bangla}	&	\textbf{Gujarati}	&	\textbf{Punjabi}	&	\textbf{Tamil}	&	\textbf{Malayalam}	&	\textbf{Malayalam}	&	\textbf{Kannada}	\\ 
    & & & & & & \textbf{intra-word} & & & & & & & & \textbf{scripts} \\ \midrule
	&		&	111	&	1	&	1	&	12	&	1	&	10	&	7	&	1	&	1	&	2	&	37	&	23	&	1	&	10	\\ \midrule
\cite{joshi1982processing}	&	COLING	&		&	\cellcolor{blue!25}$\checkmark$	&		&		&		&		&		&		&		&		&		&		&		&		\\
\cite{sankoff1998production}	&	COLING	&		&		&		&		&		&		&		&		&		&		&	\cellcolor{blue!25}$\checkmark$	&		&		&		\\
\cite{bhattacharja2010benglish}	&	PACLIC	&		&		&		&		&	\cellcolor{blue!25}$\checkmark$	&		&		&		&		&		&		&		&		&		\\
\cite{diab2011feasibility}	&	ALR &	\cellcolor{blue!25}$\checkmark$	&		&		&		&		&		&		&		&		&		&		&		&		&		\\
\cite{dey2014hindi}	&	LREC	&	\cellcolor{blue!25}$\checkmark$	&		&		&		&		&		&		&		&		&		&		&		&		&		\\
\bottomrule
        \end{tabular}
    }
    \caption{*CL Catalog in South Asian-English (1).}
    \label{tab:acl-catalog-south-asian-english-1}
\end{table*}

\begin{table*}[!t]
    \centering
    \resizebox{0.8\linewidth}{!}{
        \begin{tabular}{@{}lcccccccccccccccccc@{}} \toprule
\cite{wu2020meistermorxrc}	&	SemEval	&	\cellcolor{blue!25}$\checkmark$	&		&		&		&		&		&		&		&		&		&		&		&		&		\\
\cite{baroi2020nits}	&	SemEval	&	\cellcolor{blue!25}$\checkmark$	&		&		&		&		&		&		&		&		&		&		&		&		&		\\
\cite{gopalan2020reed}	&	SemEval	&	\cellcolor{blue!25}$\checkmark$	&		&		&		&		&		&		&		&		&		&		&		&		&		\\
\cite{malte2020team_swift}	&	SemEval	&	\cellcolor{blue!25}$\checkmark$	&		&		&		&		&		&		&		&		&		&		&		&		&		\\
\cite{zaharia2020upb}	&	SemEval	&	\cellcolor{blue!25}$\checkmark$	&		&		&		&		&		&		&		&		&		&		&		&		&		\\
\cite{zhu2020zyy1510}	&	SemEval	&	\cellcolor{blue!25}$\checkmark$	&		&		&		&		&		&		&		&		&		&		&		&		&		\\
\cite{parekh2020understanding}	&	CoNLL	&	\cellcolor{blue!25}$\checkmark$	&		&		&		&		&		&		&		&		&		&		&		&		&		\\
\cite{chakravarthi2020sentiment}	&	SLTU \& CCURL	&		&		&		&		&		&		&		&		&		&		&		&	\cellcolor{blue!25}$\checkmark$	&		&		\\
\bottomrule
        \end{tabular}
    }
    \caption{*CL Catalog in South Asian-English (2).}
    \label{tab:acl-catalog-south-asian-english-2}
\end{table*}

\begin{table*}[!t]
    \centering
    \resizebox{0.83\linewidth}{!}{
        \begin{tabular}{@{}lccccc@{}} \toprule
	\textbf{Paper} & \textbf{Proceeding} &	\textbf{Vietnamese}	&	\textbf{Tagalog}	&	\textbf{Indonesian}	\\ \midrule
	&		&	1	& 2 &	2	\\ \midrule
\cite{oco2012pattern} &	PACLIC	&		&	\cellcolor{blue!25}$\checkmark$	&			\\ 
\cite{stymne2020evaluating} &	CALCS	&		&		&		\cellcolor{blue!25}$\checkmark$	\\
\cite{nguyen2020canvec} &	LREC	&	\cellcolor{blue!25}$\checkmark$	&		&			\\
\cite{arianto2020aspect} &	PACLIC	&		&		&		\cellcolor{blue!25}$\checkmark$	\\
\cite{herrera2022tweettaglish} &	LREC	&		&		&	\cellcolor{blue!25}$\checkmark$		\\
\bottomrule
        \end{tabular}
    }
    \caption{*CL Catalog in South East Asian-English.}
    \label{tab:acl-catalog-sea-english}
\end{table*}

\begin{table*}[!t]
    \centering
    \resizebox{\linewidth}{!}{
        \begin{tabular}{@{}lcccccccc@{}} \toprule
	\textbf{Paper} &	\textbf{Proceeding}	&	\textbf{Darija-MSA} &	\textbf{MSA-Egyptian} &	\textbf{MSA-Other Dialect}	&	\textbf{Chinese-Taiwanese}	&	\textbf{MSA-Levant Arabic}	&	\textbf{MSA-Gulf}	&	\textbf{Mixed-English}	\\ \midrule
	&		&	1	&	15	&	10	&	2	&	2	&	1	&	1	\\ \midrule
\cite{chu2007language}	&		&		&		&	\cellcolor{blue!25}$\checkmark$	&		&		&		\\
\cite{yu2012language}	&	CIPS-SIGHAN	&		&		&		&	\cellcolor{blue!25}$\checkmark$	&		&		&		\\
\cite{elfardy2012token} &	COLING	&		&	\cellcolor{blue!25}$\checkmark$	&		&		&	\cellcolor{blue!25}$\checkmark$	&		&		\\
\cite{solorio2014overview}	&	CALCS	&		&		&	\cellcolor{blue!25}$\checkmark$	&		&		&		&		\\
\cite{chittaranjan2014word}	&	CALCS	&		&		&	\cellcolor{blue!25}$\checkmark$	&		&		&		&		\\
\cite{lin2014cmu}	&	CALCS	&		&		&	\cellcolor{blue!25}$\checkmark$	&		&		&		&		\\
\bottomrule
        \end{tabular}
    }
    \caption{*CL Catalog in Dialect.}
    \label{tab:acl-catalog-dialect}
\end{table*}

\begin{table*}[!t]
    \centering
    \resizebox{\linewidth}{!}{
        \begin{tabular}{@{}lcccccc@{}} \toprule
	\textbf{Paper} &		\textbf{Proceeding} &	\textbf{Komi-Zyrian - Russian}	&	\textbf{Arabizi-Arabic}	&	\textbf{Spanish-Catalan}	&	\textbf{Corsican-French}	&	\textbf{Frisian-Dutch}	\\ \midrule
	&		&	1	&	1	&	1	&	1	&	3	\\ \midrule
\cite{eskander2014foreign}	&	CALCS	&		&	\cellcolor{blue!25}$\checkmark$	&		&		&		\\
\cite{yilmaz2016longitudinal}	&	LREC	&		&		&		&		&	\cellcolor{blue!25}$\checkmark$	\\
\cite{braggaar2021challenges}	&	AdaptNLP	&		&		&		&		&	\cellcolor{blue!25}$\checkmark$	\\
\cite{amin2022few}	&	BioNLP	&		&		&	\cellcolor{blue!25}$\checkmark$	&		&		\\
\cite{ozatecs2022improving}	&	Findings of NAACL	&	\cellcolor{blue!25}$\checkmark$	&		&		&		&	\cellcolor{blue!25}$\checkmark$	\\
\cite{kevers2022coswid}	&	SIGUL	&		&		&		&	\cellcolor{blue!25}$\checkmark$	&		\\
\bottomrule
        \end{tabular}
    }
    \caption{*CL Catalog in Two Languages in the same family.}
    \label{tab:acl-catalog-samefam}
\end{table*}

\begin{table*}[!t]
    \centering
    \resizebox{\linewidth}{!}{
        \begin{tabular}{@{}lcccccccccc@{}} \toprule
	\textbf{Paper} &		\textbf{Proceeding} &	\textbf{Russian-Tatar}	&	\textbf{Russian-Tatar}&	\textbf{Turkish-German	}&	\textbf{MSA-North African	}&	\textbf{French - Arabic Dialect}	&	\textbf{Dutch-Turkish}	&	\textbf{French-Algerian}	&	\textbf{Basque-Spanish	}&	\textbf{Spanish–Wixarika} \\	
&	&	&	\textbf{intra-word} & & & &	&	&	&	\textbf{intra-word} \\ \midrule									
	&		&	1	&	1	&	7	&	1	&	2	&	2	&	1	&	1	&	1	\\	\midrule									
\cite{sankoff1998production}	&	COLING	&		&		&		&		&	\cellcolor{blue!25}$\checkmark$	&		&		&		&		\\										
\cite{papalexakis2014predicting}	&	CALCS	&		&		&		&		&		&	\cellcolor{blue!25}$\checkmark$	&		&		&		\\										
\cite{gamback2016comparing}	&	LREC	&		&		&		&		&		&	\cellcolor{blue!25}$\checkmark$	&		&		&		\\										
\cite{ccetinouglu2016turkish}	&	LREC	&		&		&	\cellcolor{blue!25}$\checkmark$	&		&		&		&		&		&		\\										
\cite{ccetinouglu2016challenges}	&	CALCS	&		&		&	\cellcolor{blue!25}$\checkmark$	&		&		&		&		&		&		\\
\bottomrule
        \end{tabular}
    }
    \caption{*CL Catalog in different family.}
    \label{tab:acl-catalog-diffam}
\end{table*}

\begin{table*}[!t]
    \centering
    \resizebox{\linewidth}{!}{
        \begin{tabular}{@{}lccccccc@{}} \toprule
	
	\textbf{Paper} &	\textbf{Proceeding}	&	\textbf{Tulu-Kannada-EN}	&	\textbf{Hindi-Bengali-EN}	&	\textbf{Greek-German-EN}	&	\textbf{Magahi-Hindi-EN}	&	\textbf{Arabic-EN-French}	&	\textbf{Darija-EN-French}	\\									
 \midrule
	&		&	1	&	1	&	1	&	1	&	1	&	1	\\	
 \midrule
\cite{voss2014finding} 	&	LREC	&		&		&		&		&		&	\cellcolor{blue!25}$\checkmark$	\\																
\cite{ccetinouglu2016challenges}	&	CALCS	&		&		&	\cellcolor{blue!25}$\checkmark$	&		&		&		\\																
\cite{barman2016part} &	CALCS	&		&	\cellcolor{blue!25}$\checkmark$	&		&		&		&		\\																
\cite{abdul2020toward}	&	EMNLP	&		&		&		&		&	\cellcolor{blue!25}$\checkmark$	&		\\																
\cite{taguchi2021transliteration}	&	CALCS	&		&		&		&		&		&		\\																
\cite{rani2022mhe}	&	LREC	&		&		&		&	\cellcolor{blue!25}$\checkmark$	&		&		\\																
\cite{hegde2022corpus}	&	ELRA	&	\cellcolor{blue!25}$\checkmark$	&		&		&		&		&		\\																
\bottomrule
        \end{tabular}
    }
    \caption{*CL Catalog in Trilingual.}
    \label{tab:acl-catalog-3lang}
\end{table*}

\begin{table*}[!t]
    \centering
    \resizebox{\linewidth}{!}{
        \begin{tabular}{@{}lcccccccccc@{}} \toprule						\textbf{Paper}							&	\textbf{Proceeding}	&	\textbf{SEA Mandarin-English} &	\textbf{Bangla-Chinese-Dutch-}	&	\textbf{Early New High German,} & \textbf{Telugu,}&	\textbf{MSA, Berber, French,} &	\textbf{Others (4+)}	&	\textbf{English, Swiss German} &	\textbf{Algerian, MSA,}	&	\textbf{Mandarin-Hakka-} \\	
        & & & \textbf{English-Farsi-German-} & \textbf{Latin, French, Greek,} & \textbf{Modern Standard Telugu,} & \textbf{local Algerian Arabic} & & \textbf{Latin} & \textbf{local Arabic varieties,} & \textbf{Taiwanese-English}
\\
 & & & \textbf{Hindi-Korean-Russian-} & \textbf{Italian, Hebrew} & \textbf{English, Hindi, Urdu, }& & & & \textbf{Berber, French, and English} \\ 
 & & & \textbf{Spanish-Turkish} \\ \midrule
	&		&	10	&	4	&	1	&	1	&	2	&	3	&	1	&	1	&	1	\\			\midrule							
\cite{wu2011unknown}	&	ROCLING	&		&		&		&		&		&		&		&		&	\cellcolor{blue!25}$\checkmark$	\\
\cite{adel2013combination}	&	ACL	&	\cellcolor{blue!25}$\checkmark$	&		&		&		&		&		&		&		&		\\
\cite{volk2014detecting}	&	CALCS	&		&		&		&		&		&		&	\cellcolor{blue!25}$\checkmark$	&		&		\\
\bottomrule
        \end{tabular}
    }
    \caption{*CL Catalog in 4+ languages.}
    \label{tab:acl-catalog-4lang}
\end{table*}

\begin{table*}[!t]
    \centering
    \resizebox{\linewidth}{!}{
        \begin{tabular}{@{}lcccccccc@{}} \toprule
        \textbf{Paper} & \textbf{Proceeding} & \textbf{isiZulu} & \textbf{isiXhosa} & \textbf{Setsawa} & \textbf{Sesotho} & \textbf{Sotho} \\ \midrule
        & & 6 & 4 & 3 & 3 & 1 \\ \midrule
        \cite{niesler2008accent} & Odyssey & \cellcolor{blue!25}$\checkmark$ & \cellcolor{blue!25}$\checkmark$ \\
        \cite{mabokela2014modeling} & SLTU & & & & & \cellcolor{blue!25}$\checkmark$ \\
        \cite{van2017synthesising} & Interspeech & \cellcolor{blue!25}$\checkmark$ \\
        \cite{yilmaz2018building} & Interspeech & \cellcolor{blue!25}$\checkmark$ & \cellcolor{blue!25}$\checkmark$ & \cellcolor{blue!25}$\checkmark$ & \cellcolor{blue!25}$\checkmark$ \\
        \cite{biswas2018multilingual} & Interspeech & \cellcolor{blue!25}$\checkmark$ \\
        \cite{biswas2018improving} & SLTU & \cellcolor{blue!25}$\checkmark$ & \cellcolor{blue!25}$\checkmark$ & \cellcolor{blue!25}$\checkmark$ & \cellcolor{blue!25}$\checkmark$ \\
        \cite{biswas2019semi} & Interspeech & \cellcolor{blue!25}$\checkmark$ & \cellcolor{blue!25}$\checkmark$ & \cellcolor{blue!25}$\checkmark$ & \cellcolor{blue!25}$\checkmark$ \\ \bottomrule
        \end{tabular}
    }
    \caption{ISCA Catalog in African-English.}
    \label{tab:isca-catalog-african-english}
\end{table*}

\begin{table*}[!t]
    \centering
    \resizebox{0.87\linewidth}{!}{
        \begin{tabular}{@{}lccccccc@{}} \toprule
        \textbf{Paper} & \textbf{Proceeding} & \textbf{Chinese} & \textbf{Cantonese} & \textbf{Korean} & \textbf{Japanese} \\ \midrule
        & & 27 & 5 & 1 & 1 \\ \midrule
        \cite{fu2000model} & ISCSLP & \cellcolor{blue!25}$\checkmark$ \\
        \cite{kojima2003mixed} & Eurospeech & & & & \cellcolor{blue!25}$\checkmark$ \\
        \cite{you2004chinese} & ISCSLP & \cellcolor{blue!25}$\checkmark$ \\
        \cite{chan2004detection} & ISCSLP & & \cellcolor{blue!25}$\checkmark$ \\
        \cite{chan2005development} & Interspeech & & \cellcolor{blue!25}$\checkmark$ \\
        \cite{ren2005mandarin} & Interspeech & \cellcolor{blue!25}$\checkmark$ \\
\bottomrule
        \end{tabular}
    }
    \caption{ISCA Catalog in East Asian-English.}
    \label{tab:isca-catalog-eastasian-english}
\end{table*}

\begin{table*}[!t]
    \centering
    \resizebox{\linewidth}{!}{
        \begin{tabular}{@{}lccccccc@{}} \toprule
        \textbf{Paper} & \textbf{Proceeding} & \textbf{Spanish} & \textbf{French} & \textbf{German} & \textbf{Maltese} \\ \midrule
        \cite{pfister03b_eurospeech} & Eurospeech & & & \cellcolor{blue!25}$\checkmark$ \\
        \cite{romsdorfer05_interspeech} & Interspeech & & \cellcolor{blue!25}$\checkmark$ \\
        \cite{rosner07_interspeech} & Interspeech & & & & \cellcolor{blue!25}$\checkmark$ \\
        \cite{piccinini14_speechprosody} & SpeechProsody & \cellcolor{blue!25}$\checkmark$ \\
        \cite{sitaram16_ssw} & SSW & & & \cellcolor{blue!25}$\checkmark$ \\
        \cite{soto17_interspeech} & Interspeech & \cellcolor{blue!25}$\checkmark$ \\
        \cite{ramanarayanan17_interspeech} & Interspeech & \cellcolor{blue!25}$\checkmark$ \\
        \cite{guzman17_interspeech} & Interspeech & \cellcolor{blue!25}$\checkmark$ \\
        \cite{bullock18_interspeech} & Interspeech & \cellcolor{blue!25}$\checkmark$ \\
        \cite{soto18_interspeech} & Interspeech & \cellcolor{blue!25}$\checkmark$ \\
        \cite{soto19_interspeech} & Interspeech & \cellcolor{blue!25}$\checkmark$ \\
        \cite{chandu20_interspeech} & Interspeech & \cellcolor{blue!25}$\checkmark$ \\ \bottomrule
        \end{tabular}
    }
    \caption{ISCA Catalog in European-English.}
    \label{tab:isca-catalog-european-english}
\end{table*}

\begin{table*}[!t]
    \centering
    \resizebox{0.65\linewidth}{!}{
        \begin{tabular}{@{}lccccc@{}} \toprule
        \textbf{Paper} & \textbf{Proceeding} & \textbf{Modern Standard Arabic} \\ \midrule
        \cite{white2008investigation} & Interspeech & \cellcolor{blue!25}$\checkmark$ \\
        \cite{ali21b_interspeech} & Interspeech & \cellcolor{blue!25}$\checkmark$ \\
        \cite{chowdhury21_interspeech} & Interspeech & \cellcolor{blue!25}$\checkmark$ \\ \bottomrule
        \end{tabular}
    }
    \caption{ISCA Catalog in Middle Eastern-English.}
    \label{tab:isca-catalog-middleeastern-english}
\end{table*}

\begin{table*}[!t]
    \centering
    \resizebox{\linewidth}{!}{
        \begin{tabular}{@{}lccccccccccc@{}} \toprule
        \textbf{Paper} & \textbf{Proceeding} & \textbf{Hindi} & \textbf{Marathi} & \textbf{Bengali} & \textbf{Telugu} & \textbf{Gujarati} & \textbf{Tamil} & \textbf{Malayalam} & \textbf{Kannada} \\ \midrule
        \cite{sitaram16_ssw} & SSW & \cellcolor{blue!25}$\checkmark$ \\
        \cite{ramanarayanan17_interspeech} & Interspeech & \cellcolor{blue!25}$\checkmark$ \\
        \cite{ganji18_interspeech} & Interspeech & \cellcolor{blue!25}$\checkmark$ \\
        \cite{rao18_interspeech} & Interspeech & \cellcolor{blue!25}$\checkmark$ \\
        \cite{thomas18_interspeech} & Interspeech & \cellcolor{blue!25}$\checkmark$ & & & & & \cellcolor{blue!25}$\checkmark$ \\
        \cite{srivastava18_interspeech} & Interspeech & \cellcolor{blue!25}$\checkmark$ \\
        \cite{rambabu18_sltu} & SLTU & \cellcolor{blue!25}$\checkmark$ \\
\bottomrule
        \end{tabular}
    }
    \caption{ISCA Catalog in South Asian-English.}
    \label{tab:isca-catalog-southasian-english}
\end{table*}

\begin{table*}[!t]
    \centering
    \resizebox{0.5\linewidth}{!}{
        \begin{tabular}{@{}lccccc@{}} \toprule
        \textbf{Paper} & \textbf{Proceeding} & \textbf{Malay} \\ \midrule
        \cite{yeong10_sltu} & SLTU & \cellcolor{blue!25}$\checkmark$ \\
        \cite{yeong14_interspeech} & Interspeech & \cellcolor{blue!25}$\checkmark$ \\
        \cite{singh18_sltu} & Interspeech & \cellcolor{blue!25}$\checkmark$
        \\ \bottomrule
        \end{tabular}
    }
    \caption{ISCA Catalog in South East Asian-English.}
    \label{tab:isca-catalog-southeastasian-english}
\end{table*}

\begin{table*}[!t]
    \centering
    \resizebox{0.6\linewidth}{!}{
        \begin{tabular}{@{}lccccc@{}} \toprule
        \textbf{Paper} & \textbf{Proceeding} & \textbf{Chinese-Taiwanese} \\ \midrule
        \cite{lyu08_interspeech} & Interspeech & \cellcolor{blue!25}$\checkmark$
        \\ \bottomrule
        \end{tabular}
    }
    \caption{ISCA Catalog in Language with Dialects.}
    \label{tab:isca-catalog-language-with-dialects}
\end{table*}

\begin{table*}[!t]
    \centering
    \resizebox{0.8\linewidth}{!}{
        \begin{tabular}{@{}lccccc@{}} \toprule
        \textbf{Paper} & \textbf{Proceeding} & \textbf{Frisian-Dutch} & \textbf{Russian-Ukrainan} \\ \midrule
        \cite{lyudovyk14_sltu} & Interspeech & & \cellcolor{blue!25}$\checkmark$ \\ 
        \cite{ylmaz16b_interspeech} & Interspeech & \cellcolor{blue!25}$\checkmark$ \\  
        \cite{ylmaz17b_interspeech} & Interspeech & \cellcolor{blue!25}$\checkmark$ \\  
        \cite{ylmaz17_interspeech} & Interspeech & \cellcolor{blue!25}$\checkmark$ \\  
        \cite{ylmaz18_interspeech} & Interspeech & \cellcolor{blue!25}$\checkmark$ \\  
        \cite{yilmaz18_sltu} & SLTU & \cellcolor{blue!25}$\checkmark$ \\  
        \cite{wang19l_interspeech} & Interspeech & & \cellcolor{blue!25}$\checkmark$ \\  
        \cite{ylmaz19b_interspeech} & Interspeech & & \cellcolor{blue!25}$\checkmark$ 
        \\ \bottomrule
        \end{tabular}
    }
    \caption{ISCA Catalog in Two Languages in the same family.}
    \label{tab:isca-catalog-two-langauges-in-the-same-family}
\end{table*}

\begin{table*}[!t]
    \centering
    \resizebox{0.83\linewidth}{!}{
        \begin{tabular}{@{}lccccc@{}} \toprule
        \textbf{Paper} & \textbf{Proceeding} & \textbf{Kazakh-Russian} & \textbf{Hindi-Tamil} & \textbf{French-Arabic} & \\ \midrule
        & & 1 & 1 & 4 \\ \midrule
        \cite{amazouz17_interspeech} & Interspeech & & & \cellcolor{blue!25}$\checkmark$ \\
        \cite{thomas18_interspeech} & Interspeech & & \cellcolor{blue!25}$\checkmark$ \\
        \cite{wottawa18_interspeech} & Interspeech & & & \cellcolor{blue!25}$\checkmark$ \\
        \cite{chandu20_interspeech} 
        & Interspeech & & & \cellcolor{blue!25}$\checkmark$ \\ 
        \cite{chowdhury21_interspeech} & Interspeech & & & \cellcolor{blue!25}$\checkmark$ \\
        \cite{mussakhojayeva22_interspeech} & Interspeech & \cellcolor{blue!25}$\checkmark$ 
        \\ \bottomrule
        \end{tabular}
    }
    \caption{ISCA Catalog in Two Languages in different families.}
    \label{tab:isca-catalog-two-langauges-in-different-families}
\end{table*}

\begin{table*}[!t]
    \centering
    \resizebox{0.83\linewidth}{!}{
        \begin{tabular}{@{}lccccc@{}} \toprule
        \textbf{Paper} & \textbf{Proceeding} & \textbf{Italian-German-English} & \textbf{ Kiswahili-Shen-English} \\ \midrule
        & & 1 & 1 \\ \midrule
        \cite{knill20_interspeech} & Interspeech & \cellcolor{blue!25}$\checkmark$ \\
        \cite{otundo22_speechprosody} & SpeechProsody & & \cellcolor{blue!25}$\checkmark$
        \\ \bottomrule
        \end{tabular}
    }
    \caption{ISCA Catalog in Trilingual.}
    \label{tab:isca-catalog-trilingual}
\end{table*}

\begin{table*}[!t]
    \centering
    \resizebox{\linewidth}{!}{
        \begin{tabular}{@{}lccccc@{}} \toprule
        \textbf{Paper} & \textbf{Proceeding} & \textbf{SEA Mandarin-English} & \textbf{African Languages-English} & \textbf{Indian Languages-English} & \textbf{Others} \\ \midrule
        & & 17 & 1 & 1 & 7 \\ \midrule
        \cite{badino04b_interspeech} & Interspeech & & & & \cellcolor{blue!25}$\checkmark$\\
        \cite{oria04_interspeech} & Interspeech & & & &  \cellcolor{blue!25}$\checkmark$ \\
        \cite{marcadet05_interspeech} & Interspeech & & & & \cellcolor{blue!25}$\checkmark$ \\
        \cite{romsdorfer06_ml} & ML  & & & & \cellcolor{blue!25}$\checkmark$ \\
        \cite{lyu10_interspeech} & Interspeech & \cellcolor{blue!25}$\checkmark$ \\
        \cite{imseng10_interspeech} & Interspeech & & & & \cellcolor{blue!25}$\checkmark$ \\
\bottomrule
        \end{tabular}
    }
    \caption{ISCA Catalog in 4+.}
    \label{tab:isca-catalog-4+}
\end{table*}

% \begin{table*}[!t]
%     \centering
%     \resizebox{\linewidth}{!}{
%         \begin{tabular}{@{}lcccc@{}} \toprule
%         \textbf{Paper} & \textbf{Language Category} & \textbf{Proceeding} & \textbf{Proceeding Type} &  \\ \midrule
%         \citet{traber1999multilingual} & NA & Eurospeech & Conference \\
%         \citet{fu2000model} & Chinese-English & ISCSLP & Symposium \\ 
%         \citet{kojima2003mixed} & East Asian-English & Eurospeech & Conference \\
%         \citet{pfister2003mixed} & European-English & Eurospeech & Conference \\
%         \citet{badino2004general} & 4+ & Interspeech & Conference \\
%         \citet{oria2004multilingual} & 4+ & Interspeech & Conference \\
%         \citet{you2004chinese} & East Asian-English & ISCSLP & Symposium \\
%         \citet{chan2004detection} & East Asian-English & ISCSLP & Symposium \\
%         \citet{marcadet2005transformation} & & & \\
%         \\ \bottomrule
%         \end{tabular}
%     }
%     \caption{ISCA Catalog in East Asian-English.}
%     \label{tab:isca-catalog-eastasian-english}
% \end{table*}

\end{document}